\pdfoutput=1
\documentclass[runningheads]{llncs}
\usepackage{graphicx}

\usepackage{tikz}
\usepackage{comment}
\usepackage{amsmath,amssymb} 
\usepackage{color}
\usepackage[pagebackref,breaklinks,colorlinks]{hyperref}
\usepackage[capitalize]{cleveref}

\usepackage[accsupp]{axessibility}  

\usepackage{multirow}
\usepackage{graphicx}
\usepackage{amsmath}
\usepackage{amssymb}
\usepackage{booktabs}

\usepackage{tabularx}
\usepackage{bm}
\usepackage{wrapfig}
\usepackage{subfigure}
\usepackage{caption}

\begin{document}
\pagestyle{headings}
\mainmatter
\def\ECCVSubNumber{5961}  

\title{Global Spectral Filter Memory Network for Video Object Segmentation} 


\titlerunning{Global Spectral Filter Memory Network for Video Object Segmentation}
%

\author{Yong Liu\inst{1*} \and
Ran Yu\inst{1} \and
Jiahao Wang\inst{1} \and
Xinyuan Zhao\inst{3} \and
Yitong Wang\inst{2} \and
Yansong Tang\inst{1} \and
Yujiu Yang\inst{1}$^{\dagger}$
}
\authorrunning{Y. Liu et al.}
%
\institute{Tsinghua Shenzhen International Graduate School, Tsinghua University \and
ByteDance Inc. \and
Northwestern University\\
\email{\{liu-yong20, yu-r19\}@mails.tsinghua.edu.cn, \{tang.yansong, yang.yujiu\}@sz.tsinghua.edu.cn}}
\maketitle
\let\thefootnote\relax\footnotetext{\scriptsize{$^*$This work was done during an internship at ByteDance Inc.}}
\let\thefootnote\relax\footnotetext{\scriptsize{$^{\dagger}$Corresponding author}}

\begin{abstract}
This paper studies semi-supervised video object segmentation through boosting intra-frame interaction. Recent memory network-based methods focus on exploiting inter-frame temporal reference while paying little attention to intra-frame spatial dependency. Specifically, these segmentation model tends to be susceptible to interference from unrelated nontarget objects in a certain frame. To this end, we propose Global Spectral Filter Memory network (GSFM), which improves intra-frame interaction through learning long-term spatial dependencies in the spectral domain. The key components of GSFM is 2D (inverse) discrete Fourier transform for spatial information mixing. Besides, we empirically find low frequency feature should be enhanced in encoder (backbone) while high frequency for decoder (segmentation head). We attribute this to semantic information extracting role for encoder and fine-grained details highlighting role for decoder. Thus, Low (High) Frequency Module is proposed to fit this circumstance. 
Extensive experiments on the popular DAVIS and YouTube-VOS benchmarks
demonstrate that GSFM noticeably outperforms the baseline method and achieves state-of-the-art performance. Besides, extensive analysis shows that the proposed modules are reasonable and of great generalization ability.
Our source code is available at \textcolor{magenta}{\url{https://github.com/workforai/GSFM}}.

\end{abstract}

\section{Introduction}

Video Object Segmentation (VOS)~\cite{davis16,davis17,youtube,hin} aims at identifying and segmenting objects in videos. It is one of the most challenging tasks in computer vision 
with many potential applications, including interactive video editing, augmented reality~\cite{application1}, and autonomous driving~\cite{application2}.
In this paper, we focus on the semi-supervised setting where target objects are defined by the given masks of the first frame.
It is crucial for semi-supervised VOS to fully utilize the available reference information to distinguish targets from background objects.


Since the critical problem of this task lies in how to make full use of the spatial-temporal dependency to recognize the targets,
matching-based methods, which perform pixel-level matching with historical reference frames, have received tremendous attention.
The Space-Time Memory Network~\cite{stm} memorizes intermediate frames with segmentation masks as references and performs pixel-level matching between them with the current frame to segment target objects in a bottom-up manner, which has been proved effective and has served as the current mainstream framework.
Some works~\cite{kmn,gcm,mivos,lcm,rmnet,hmm,swiftnet,stcn,aaai,break_shortcut,afb-urr,QDMN} further develop STM and have achieved excellent performance.

\begin{figure*}[t]
    \centering
    \includegraphics[width=\textwidth]{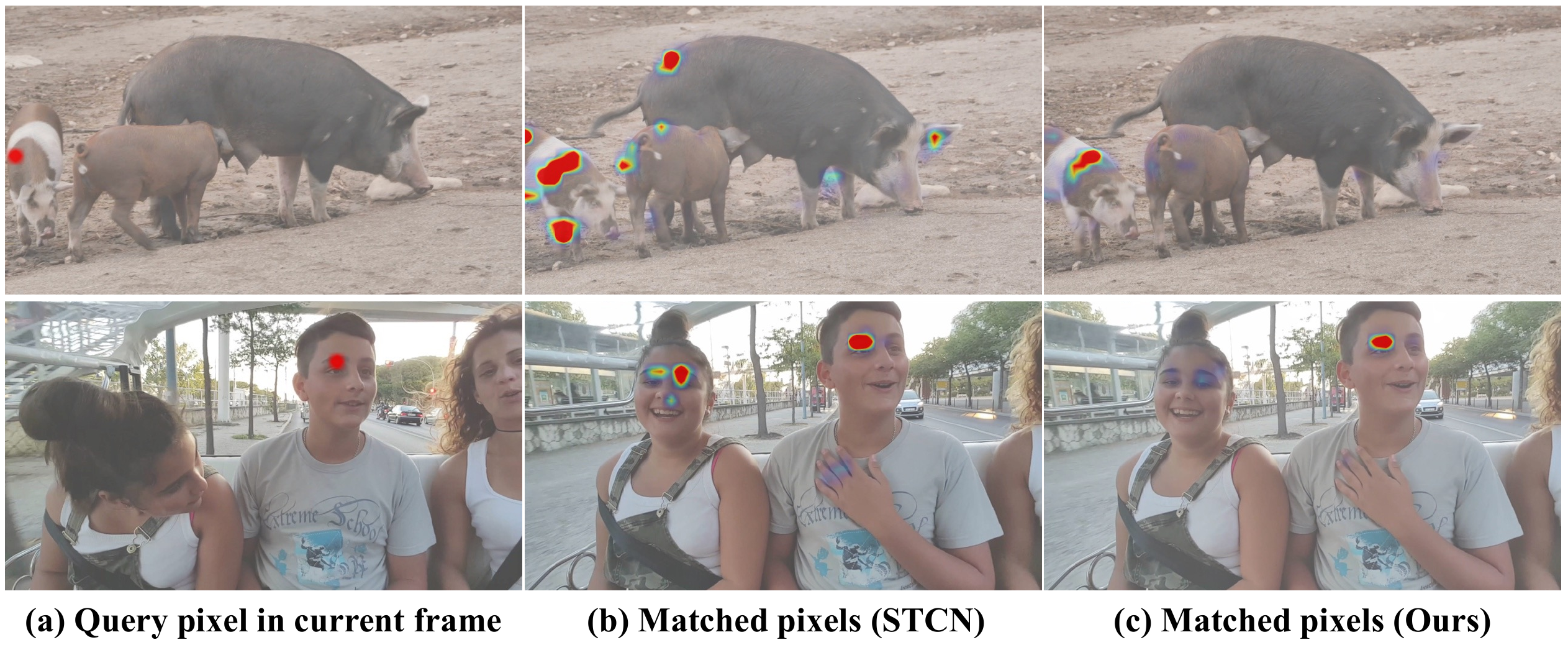}
    \caption{Illustration of the disadvantages of lacking semantic global information. The highlight red pixels in the first column are target pixels. The second column shows that previous method~\cite{stcn} would incorrectly match similar pixels. In the third column, our model relieves the confusion problem by enhancing low-frequency components and updating features from spectral domain.}
    \label{intro}
\end{figure*}

Although these methods have made great progress in the field of VOS, they pay little attention to excavating intra-frame dependency and only utilize local representation for matching and prediction due to the inductive bias of convolution.
Lacking global dependency would cause low efficacy in distinguishing similar pixels, \textit{e.g.}, pixels of similar color or objects of the same category.
We take the typical method STCN~\cite{stcn} for illustration. 
In~\cref{intro} (b), some pixels belonging to background objects are mismatched with the target pixel due to their similar local features.
Ignoring long-range dependency for matching would lead to a high risk of interference from other objects.
Since the matching-based approaches rely on the matching process to identify the targets, incorrectly matched pixels would negatively affect the final segmentation and even lead to error accumulation.
Therefore, it is necessary to excavate the intra-frame spatial dependency to enhance the representation of features.

According to the Fourier theory~\cite{convolution_theorem}, FFT function generates outputs based on pixels from all spatial locations when processing input feature.
Thus, the spectral domain representation contains rich global information.
Inspired by this, we introduce a Global Spectral Filter Memory network (GSFM), which fuses global dependency from spectral domain and distinguishes the high-frequency and low-frequency components for targeted enhancement.
In GSFM, we propose the Low Frequency Module (LFM) and High Frequency Module (HFM) to enhance different representation according to the characteristics of the encoder-decoder network structure.

The role of encoder is to extract deep features for subsequent modules, and the encoded features need to contain rich semantic information.
Intuitively, low-frequency components correspond to high-level semantic information while ignoring details.
Some theoretical researches on CNN from spectral domain~\cite{frequency-analysis1,frequency-analysis2,frequency-analysis3} also point out similar observations.
Inspired by the above analysis, we propose a Low-Frequency Module (LFM) for the encoding process to update the features in the spectral domain and emphasize their low-frequency components.
\cref{intro} (c) illustrates that with LFM enhancing global semantic information, the distinguishability of similar pixels is greatly improved.
Extensive experiments also demonstrate the rationality of emphasizing low-frequency in the encoder.

Different from encoding, features in the decoding process need to contain more fine-grained information for accurate prediction.
And high-frequency components correspond to the image parts that change drastically, \textit{e.g.}, object boundaries and texture details.
Combined with the above analysis, we believe that focusing on high-frequency components would help to rich the fine-grained representation of features and make more accurate predictions of boundaries or ambiguous regions.
Therefore, we introduce a High-Frequency Module (HFM) in the decoding process, which enhances the high-frequency components of features to better capture detailed information.
Besides, to take full advantage of HFM, we combine it with an additional boundary prediction branch to provide better localization and shape guidance.


Experiments show that the proposed model noticeably outperforms the baseline method and achieves state-of-the-art performance on DAVIS~\cite{davis16,davis17} and YouTube-VOS~\cite{youtube} datasets.
The contribution of this paper can be summarized as follows.
Firstly, we propose to leverage the spectral domain to enhance the global spatial dependency of features for semi-supervised VOS.
Secondly, considering the differences between the process of encoding and decoding, we propose LFM and HFM to perform targeted enhancement, respectively.
Thirdly, we combine object boundaries and high-frequency to provide better localization and shape information while keeping the decoding features are fine-grained.

\section{Related Work}

\subsubsection{Semi-supervised video object segmentation.}
Since the masks for the first frame are given, early methods~\cite{osvos,onavos,onlinefitune,onlinefinetune2,monet} take the strategy that online fine-tune the network according to the object mask of the first frame, which suffers from slow inference speed.
Propagation-based methods~\cite{objectflow,segflow,favos,dvsnet,maskrnn,cinn,propagation1,propagation4,propagation5} forward propagate the segmentation masks as a reference to the next frame, and they are difficult to handle complicated scenarios.
Some other researchers have decoupled VOS into three independent subtasks of detection, tracking, and segmentation~\cite{premvos,dyenet,topdown1,topdown2}.
Although this approach balances running time and accuracy, it is extremely dependent on the performance of the detectors and makes the entire pipeline complex.

In recent years matching-based methods have received great attention for excellent performance and robustness.
FEELVOS~\cite{feelvos}, CFBI~\cite{cfbi} and CFBI$+$~\cite{cfbi+} perform global and local matching with the first frame and the previous adjacent frame, respectively.
AOT~\cite{aot} associates multiple target objects into the same embedding space by employing an identification mechanism.
STM~\cite{stm} leverages the memory network to memorize intermediate frames as references, which has been proved effective and has served as the current mainstream framework. 
Based on STM, KMN~\cite{kmn} and RMNet~\cite{rmnet} perform local-to-local matching by using the Gaussian kernel and hard crop strategy.
SwiftNet~\cite{swiftnet} and AFB-URR~\cite{afb-urr} reduce memory redundancy by calculating the similarity between query and memory.
LCM~\cite{lcm} and SCM~\cite{davis1st} proposes spatial constraint to enhance spatial location information.
EGMN~\cite{egmn} employs an episodic memory network to memorize frames as nodes and capture cross-frame correlations by edges.
MiVOS~\cite{mivos} further developed KMN~\cite{kmn} by utilizing the top-k strategy to reduce noise information in the memory read block. 
STCN~\cite{stcn} improves the feature extraction and performs more reasonable matching by decoupling the image and masks.

Despite the great performance achieved by these methods, they ignore the importance of fully excavating the intra-frame global information, which may lead to a high risk of interference by pixels with similar local features.

\subsubsection{Spectral domain learning.}
Recent years have witnessed increasing research enthusiasm on combining spectral domain and deep learning~\cite{dctmask,fast-fourier-convolution,fcanet,frequency-analysis2,frequency-analysis3,frequencyrelatedwork}.
Among them, some researches~\cite{frequency-analysis1,frequency-analysis2,frequency-analysis3} attempt to explain the behavior of convolution neural network from the perspective of spectral domain.
They point out that the features of different frequency bands represent different types of information and observe some properties of deep neural networks related to it.
With the guidance of these works and rethinking about the characteristics of the encoder-decoder structure, we propose separating the high-frequency and low-frequency components for reasonably utilizing them.
In this paper, we introduce a low-frequency module (LFM) and a high-frequency module (HFM). 
LFM enhances the low-frequency components during encoding to fuse global semantic features, while HFM enhances the high-frequency components in the decoder to make features contain more fine-grained details.

Some previous methods~\cite{afb-urr,matnet,network_modulation} applying spatial prior filter or introducing boundary to features can also be explained from the perspective of spectral domain.
Applying filter kernels or highlighting boundaries in the spatial domain is essentially a special way to distinguish between high and low frequencies. 
While this approach can also serve the purpose of targeted enhancement, it loses the advantage of global perception in the spectral domain. 
Therefore, our approach that updates features in the spectral domain is more generalized and effective.

\begin{figure*}[t]
    \centering
    \includegraphics[width=\textwidth]{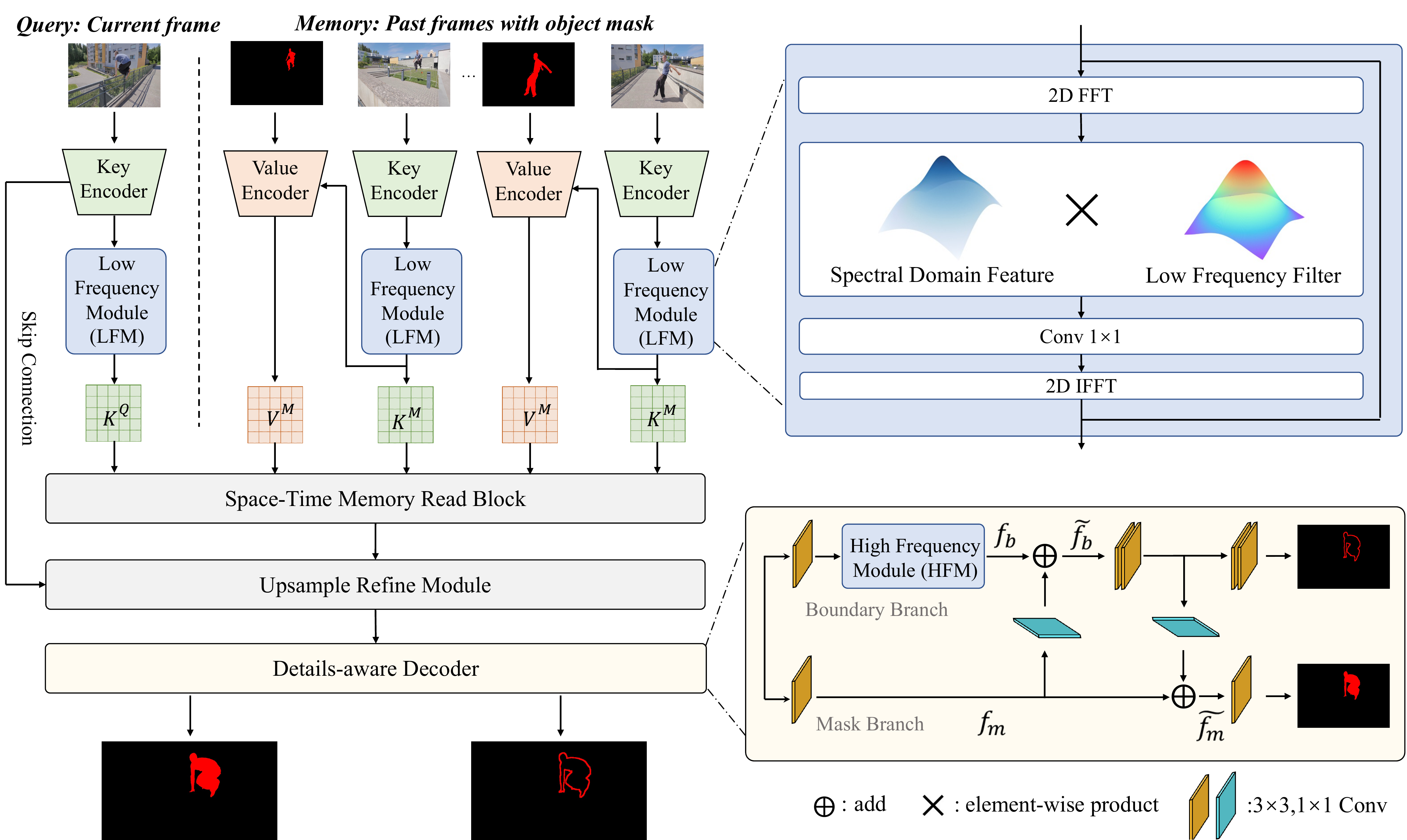}
    \caption{Overview of GSFM. The network takes both query (current frame) and memory (past frames with masks) as input. LFM enhances low-frequency components of features and fuses global information in the spectral domain.
    Having $K^M$, $K^Q$ and $V^M$ extracted from the encoder, the memory read block calculates  similarity between query and memory.
    The refine module upsamples the features and outputs to the decoder.
    With HFM enhancing high-frequency components, the decoder jointly predicts object masks and boundaries.
    }
    \label{framework}
\end{figure*}

\section{Method}

\subsection{Overview}
The overall architecture of our GSFM is shown in~\cref{framework}.
Given a video sequence and the annotation of the first frame, we process it frame by frame.
During processing, the current frame is considered a query, and the past reference frames with segmentation masks are memory.
Following the baseline method STCN~\cite{stcn}, a Key Encoder extracts key features for each frame, and a Value Encoder extracts value features only for memory frames.
By performing matching between query and memory in Space-Time Memory Read Block, the decoder identifies and segments the target object in a bottom-up manner.
In the encoder, for exploiting the intra-frame semantic information to improve the representative capacity of features and promote the effectiveness of matching, a low-frequency module (LFM) enhances the low-frequency components of the features and performs global information updating from the spectral domain.
In the decoder, the high-frequency module (HFM) enhances high-frequency components to highlight fine-grained information for accurate prediction.
Besides, we take the strategy that jointly learning object boundaries and masks
in an end-to-end manner~\cite{boundary2,boundary3,bmaskrcnn,boundary4}.
With the interaction between mask branch and boundary branch features, the network can better perceive the localization and shape information, which also helps identify the target objects.


\subsection{Frequency Modules}
\begin{wrapfigure}{r}{5.5cm}
		\begin{minipage}[h]{1.0\linewidth}
            \centering
			\includegraphics[width=0.9\linewidth]{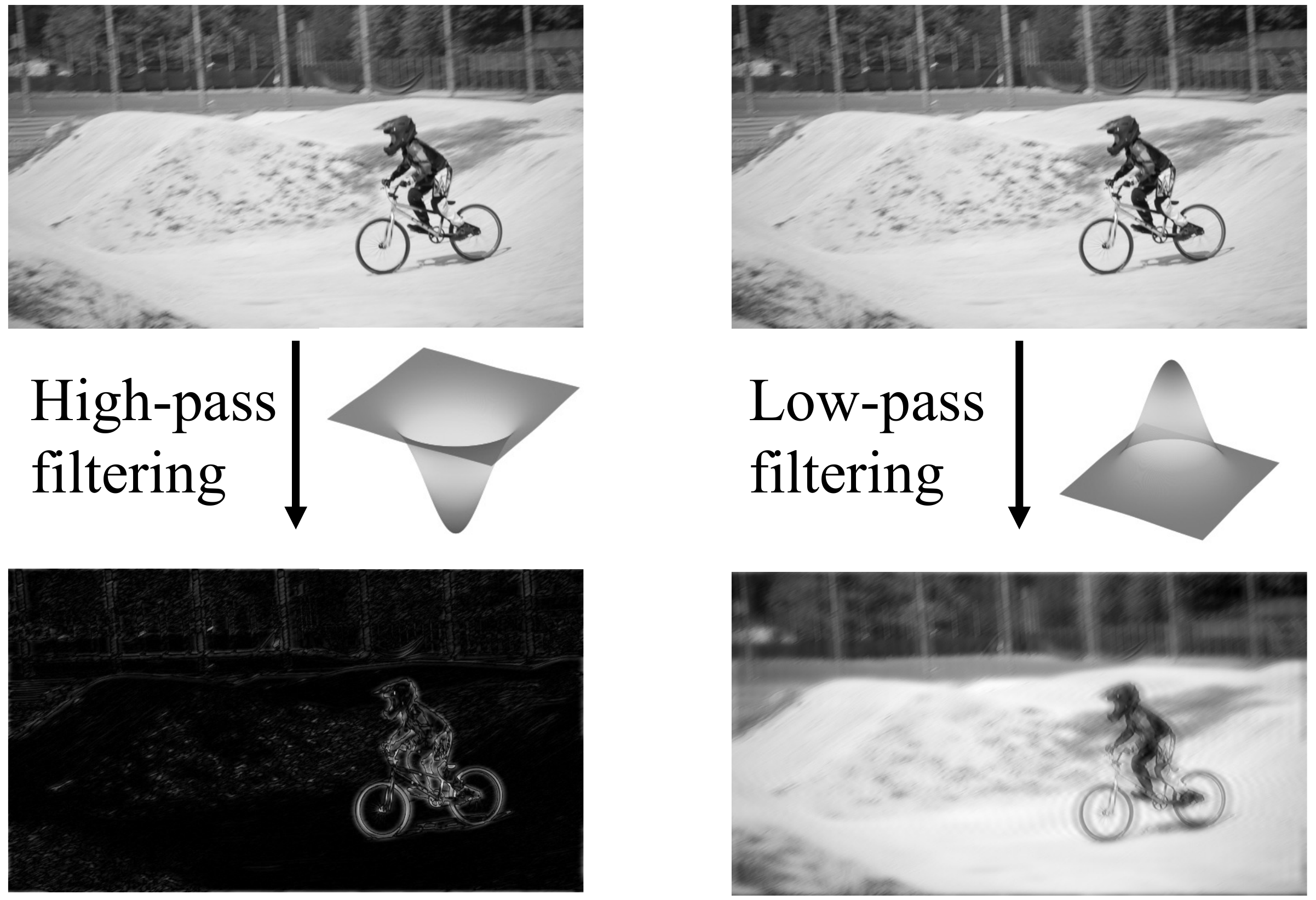}
			\caption{Illustration of different frequency components. The top line is the original image.}
			\label{high-low-component}
		\end{minipage}
	\end{wrapfigure}

According to the spectral convolution theorem~\cite{fft_theory} in Fourier theory, updating a single value in the spectral domain affects globally all original data, which sheds light on design operations with the non-local receptive field.
Intuitively, the high-frequency components correspond to the pixels varying drastically, such as object boundaries and textures, while the low-frequency components correspond to the general semantic information.
Some previous theoretical studies~\cite{frequency-analysis1,frequency-analysis2,frequency-analysis3} on spectral-domain and deep learning also point similar observation.
Besides,
to show the information represented by different frequency components more vividly, we take~\cref{high-low-component} as an example (for convenience, we use the grayscale 
image).
In~\cref{high-low-component}, the remaining information after high-pass filtering is the edges and details of objects.
After low-pass filtering, the result is an image that retains the general semantic information (some details and noise are blurred).
Considering that the role of the encoder is to extract high-level semantic information, while the decoder pursues focus on detailed features, we believe that this difference is similar to the difference between the high and low-frequency components.
Thus, we propose a low-frequency module (LFM) for the encoder and a high-frequency module (HFM) for the decoder.
\begin{figure}[t]
    \centering
    \includegraphics[width=1\textwidth]{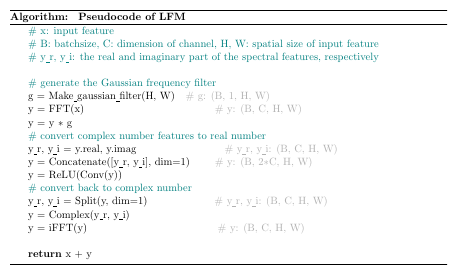}
    \caption{The Pseudocode of Low Frequency Module (LFM)}
    \label{fm}
\end{figure}

The architecture of LFM and HFM is the same, and their difference is the frequency domain filter (LFM is a low-pass filter, and HFM is a high-pass filter).
In our experiments, the filter is set in the form of Gaussian.
Here we take LFM as an example to introduce the process.
As shown in~\cref{fm},
having the image feature tensor $x$, LFM first transfers it to the spectral domain by FFT.
Then the spectral features $y$ will be passed through a low-pass filter to enhance low-frequency components, which helps to make features rich in global semantic information.
Specifically, we generate a coefficient map $g$ with the same spatial size of the feature $y$ and perform element-wise multiplication between them with the help of broadcast mechanism.
For LFM, the center of the coefficient map has the value of 0 and increases around in the form of Gaussian (without spectrum centralization, the center of the spectrum after FFT is high frequency, and the surrounding is low frequency).
Before updating the spectral domain, note that the spectral features are complex numbers for the FFT operation.
To make the complex number features compatible with the neural layers, we split the complex number into a real part $y\underline{\ }r$ and an imaginary part $y\underline{\ }i$.
For ease of computation, we append the imaginary part to the real part by concatenating them along the channel dimension, forming a new tensor with double channels.
Essentially, the resultant tensor is treated as a vanilla real number tensor, and we can perform a series of neural layers on it.
To update features in the spectral domain, we utilize 1$\times$1 convolution with ReLU activation function.
According to the convolution theorem~\cite{convolution_theorem}, convolution in one domain equals point-wise multiplication in the other domain, which implies that the 1$\times$1 convolution in spectral-domain incurs a global update in the spatial domain. 
After that, the results are converted back to complex numbers by splitting them into real and imaginary parts along the channel dimension.
Inverse 2-D FFT operation transfers the spectral features back to the spatial domain.
Finally, LFM outputs the enhanced features by adding the updated features $y$ with initial tensor $x$.

\subsection{Details-aware Decoder}
Only utilizing local information for pixel-level mask prediction may lead to a lack of overall perception of the objects and an over-reliance on pixel appearance information such as pixel color.
Intuitively, object boundaries and object masks have a close relation.
It would be helpful to locate and identify target objects from the background if the model has some sense of the shape or boundary of the objects, especially with HFM highlighting detailed information.
Besides, since semi-supervised VOS is a pixel-level tracking task, accurate boundary segmentation is significant.
Otherwise, it is easy to cause error accumulation.
Therefore, we propose to combine HFM and object boundaries to provide localization and more detailed guidance.

The architecture of the decoder is shown in \cref{framework}.
Compared to the vanilla mask decoder of other memory network-based approaches, we add a branch dedicated to predicting object boundaries so that the model gives more attention to the object boundaries and shapes.
The input feature of the boundary branch is first processed by HFM to enhance its high-frequency components, which helps to better perceive fine-grained information for accurate prediction.
In addition, due to the special relationship between object boundaries and object masks, there is a lot of mutually exploitable information between their features.
Specifically, features from the mask branch can provide basic information for localizing boundaries.
After making sense of object boundaries, the shape and location information in boundary features is also conducive to guiding more precise mask predictions.
To take full advantage of the special relationship between them, we take a fusion module~\cite{bmaskrcnn} for the interaction between the mask branch and the boundary branch.
Take the Mask $\rightarrow$ Boundary (M2B) Fusion as example, the fusion process can be formulated as follows:
\begin{equation}
    \widetilde{f}_b = \mathcal{F}(f_m) + f_b,
\end{equation}
where $\widetilde{f}_b$ denotes the fused boundary features, $f_m$ is the mask branch feature, and $f_b$ is the boundary branch feature.
$\mathcal{F}$ is a 1 $\times$ 1 convolution with ReLU function.
The fusion block is the same for the boundary $\rightarrow$ Mask (M2B) Fusion.\\

\noindent {\textbf{Boundary Ground Truth.}}
Following previous
 works~\cite{edge_detection1,edge_detection2}, we take the
boundary prediction as a pixel-level classification problem.
Since only the ground truth of the mask is available in the video object segmentation dataset,
we use the Laplacian operator to generate the boundary ground truth.
The Laplacian operator is a second-order gradient operator.
As it is regarded as a classification problem, the resultant boundaries need to be converted into binary maps, and we binarize them with a threshold of 0.1.\\

\noindent {\textbf{Boundary Loss.}}
Following previous work~\cite{bmaskrcnn}, we use dice loss~\cite{dice} and binary cross-entropy to optimize the boundary predictions.
Dice loss measures the overlap between predicted boundaries and ground truth.
More importantly, dice loss can better handle category imbalance and focus on foreground pixels, which is compatible with boundary prediction (the number of boundary points is much less than points of non-boundary).
The boundary loss $\mathcal{L}_b$ can be formulated as follows:
\begin{equation}
    \mathcal{L}_b = \mathcal{L}_{Dice} + \mathcal{L}_{BCE}.
\end{equation}
The dice loss is given as follows:
\begin{equation}
    \mathcal{L}_{Dice} = 1 - \frac{2\sum_ip^iq^i}{\sum_i(p^i)^2 + \sum_i(q^i)^2 + \epsilon},
\end{equation}
where $p$ and $q$ denote the predictions and ground truth, respectively. $i$ denotes the $i$-th pixel and $\epsilon$ is a smooth term to avoid zero division.

\subsection{Other Modules}

{\textbf{Encoder.}}
Following STCN~\cite{stcn}, we construct a Key Encoder and a Value Encoder.
For each frame, the key features are extracted only once.
In other words, we would reuse the ``query key" as the ``memory key" if one frame is memorized into the memory during video sequences.
For memory frames, since both memory keys and memory values are extracted from the same image, it is natural to reuse existing key features as the input of value encoder.
Specifically, a backbone first extracts memory features from images with segmentation masks and the resultant features are concatenated with the last layer features from key encoder.
Then two ResBlocks~\cite{resnet} and a CBAM block~\cite{cbam} process them and output the final memory value features $V^M$.\\

\noindent {\textbf{Space-Time Memory Read Block.}}
The query frame and $T$ memory frames are encoded into the followings: memory key $K^M\in\mathbb{R}^{C^k\times T\times H/16\times W/16}$, memory value ${V}^M\in\mathbb{R}^{C^v\times T\times H/16\times W/16}$, query key ${K}^Q\in\mathbb{R}^{C^k\times H/16\times W/16}$

In the Space-Time Memory Read block, activation weights are computed by measuring the similarities between $K^Q$ and $K^M$. Then the $V^M$ is retrieved by a weighted summation with the weights to get the output $M$. This operation can be summarized as:
    \begin{equation}
        M_i = \frac{1}{Z}\sum_j{\mathcal{D}(K^Q_i,K^M_j)V^M_j},
    \end{equation}
    where $i$ and $j$ are the index of the query and the memory location, $Z = \sum_j\mathcal{D}({K}^Q_i,{K}^M_j)$ is the normalizing factor. $\mathcal{D}$ denotes similarity measure (following~\cite{stcn}, in our experiments we take the L2 distance as measurement).\\

\noindent {\textbf{Refine Module.}}
We use the same refinement module as previous works~\cite{rgmp,stcn,mivos}.
The role of the refinement modules is to process the matched value features and merge the detail information from the shallow layer of the encoder.



\section{Implementation Details}
Following the training strategy in previous works~\cite{stcn,aot,mivos}, we first pretrain our model on static image datasets~\cite{static1,static2,static3,static4,static5} and then perform main training on YouTube-VOS and DAVIS datasets. 
During pretraining, each image is expanded into a pseudo video of three frames by data augmentation. 
For main training, we randomly pick three frames in chronological order (with a ground-truth mask for the first frame) from a video to form a training sample.
The range of random sampling varies with the training process.
In the intermediate period of training, the sampling range is set larger to improve the robustness of the model, while at the end of training, it is set smaller to narrow the gap between training and inference.
We use randomly cropped 384×384 patches for training.

Our models are trained with eight 32GB Tesla V100 GPUs with the Adam optimizer using PyTorch. The batch size is set to 16 for each GPU during pretraining and 8 during main training.
It takes about 18 hours to perform pretraining and 6 hours for main training.
We adopt ResNet50~\cite{resnet} as backbone for key encoders and ResNet18 for value encoder. Bootstrapped cross-entropy loss (hard example mining) is used for mask segmentation. Binary cross-entropy loss and Dice loss are used for boundary prediction.
The weight of boundary prediction loss is 0.05.
For inference, we adopt top-$k$ filtering~\cite{stcn,mivos} in our experiment with $k = 50$ in default.
We memorize every 3 frame, and no previous temporary frame is used.
Unless otherwise specified, we utilize the DAVIS2017 val set for experiment analysis.

\section{Experiments}

\begin{table}[t]
		\centering
		\caption{The quantitative evaluation on DAVIS dataset. `*' indicates our re-implementation version. The results of baseline method are underlined}
		\label{tab:results}
		\setlength\tabcolsep{3pt}
		\begin{tabular}{lcccccccccc}
			\toprule[1.5pt]
			\multirow{2}{*}{Method} & \multicolumn{3}{c}{DAVIS2016} & \multicolumn{3}{c}{DAVIS2017 val} & \multicolumn{3}{c}{DAVIS2017 test-dev} &
			\multirow{2}{*}{FPS} \\
			\cmidrule(lr){2-4} \cmidrule(lr){5-7} \cmidrule(lr){8-10}
			& $\mathcal{J}$ & $\mathcal{F}$ & $\mathcal{J}\&\mathcal{F}$ & $\mathcal{J}$ & $\mathcal{F}$ & $\mathcal{J}\&\mathcal{F}$& $\mathcal{J}$ &$\mathcal{F}$ & $\mathcal{J}\&\mathcal{F}$ \\
			\midrule
			RANet~\cite{ranet}   & 86.6 & 87.6 & 87.1 & 63.2  & 68.2 & 65.7 & 53.4  & 56.2 & 55.3 &-\\
			FEELVOS~\cite{feelvos}  & 81.1 & 82.2 & 81.7 & 69.1   & 74.0   & 71.5 & 55.2 & 60.5 & 57.8 &-\\
			RGMP~\cite{rgmp}      & 81.5   & 82.0 & 81.8 &64.8 &68.6 &66.7 &51.3 &54.4 &52.8 &-\\
			DMVOS~\cite{dmvos}   & 88.0  & 87.5  & 87.8 &- &- &- &- &- &- &-\\
			STM~\cite{stm}       & 88.7  & 89.9  & 89.3 & 79.2   & 84.3   & 81.8 & 69.3    & 75.2    & 72.2 &7.9\\
			KMN~\cite{kmn}       & 89.5  & 91.5 & 90.5 & 80.0    & 85.6    & 82.8 & 74.1    & 80.3    & 77.2 &7.1\\
			CFBI~\cite{cfbi}     & 88.3  & 90.5  & 89.4 & 79.1    & 84.6    & 81.9 & 71.1    & 78.5    & 74.8 &3.4\\
			GCM~\cite{gcm}       & 87.6  & 85.7  & 86.6 &69.3 &73.5 &71.4 &- &- &- &-\\
			G-FRTM~\cite{g-frtm} & -     & -   & 84.3  & - & - & 76.4 &- &- &- &-\\
			GIEL~\cite{giel}     &- &- &- & 80.2    & 85.3    & 82.7 & 72.0    & 78.3    & 75.2 &-\\
			SwiftNet~\cite{swiftnet} & 90.5  & 90.3  & 90.4 & 78.3    & 83.9    & 81.1 &- &- &- &20.6\\
			RMNet~\cite{rmnet}   & 88.9  & 88.7 & 88.8 & 81.0    & 86.0    & 83.5 & 71.9    & 78.1    & 75.0 &\textless11.9\\
			SSTVOS~\cite{sstvos}     &- &- &-  & 79.9  & 85.1  & 82.5   &- &- &- &-\\ 
			LCM~\cite{lcm}       & 89.9  & 91.4  & 90.7 & 80.5    & 86.5    & 83.5 & 74.4    & 81.8    & 78.1 &\textless9.5\\
			MiVOS~\cite{mivos}   & 87.8   & 90.0  & 88.9 & 80.5    & 85.8    & 83.1 &72.6    & 79.3    & 76.0 &6.5\\
			JOINT~\cite{joint}   &- &- &-  & 80.8 & 86.2  & 83.5   &- &- &-  &3.8\\
			RPCMVOS~\cite{aaai} &87.1 &94.0 &90.6 &81.3 &86.0 & 83.7 &75.8 &82.6 &79.2 &-\\
			DMN-AOA~\cite{alignment}     &- &- &- & 81.0   & 87.0 & 84.0  &74.8 &81.7 &78.3  &\textless6.2\\
			HMMN~\cite{hmm}       &89.6 &92.0 &90.8 &81.9 &87.5 &84.7  & 74.7  & 82.5  & 78.6  &6.8\\
			AOT-L~\cite{aot}  & 89.7  & 92.3 & 91.0 & 80.3   & 85.7 & 83.0 & 75.3   & 82.3 & 78.8 &15.2\\
			STCN*~\cite{stcn}    & \underline{90.1} & \underline{92.2} & \underline{91.1} & \underline{81.5} & \underline{87.9} & \underline{84.7} & \underline{72.7} & \underline{79.6} & \underline{76.1} &11.7\\
			\midrule
			\textbf{GSFM (Ours)}  & \textbf{90.1}  & \textbf{92.7} & \textbf{91.4} & \textbf{83.1}  & \textbf{89.3} & \textbf{86.2} & \textbf{74.0}  & \textbf{80.9}  & \textbf{77.5} &8.9\\
			\bottomrule[1.5pt]
		\end{tabular}
	\end{table}

\subsection{Comparisons with State-of-the-Art Methods}
    
    \textbf{DAVIS 2016}~\cite{davis16} is a densely annotated video object segmentation benchmark which contains 20 high-quality annotated video sequences.
    We compare GSFM with state-of-the-art methods in~\cref{tab:results}.
    Since the scenarios in this dataset are relatively simple and only focus on a single target object, the segmentation results of most of the methods are excellent.
    Based on the STCN~\cite{stcn}, our method achieves the performance of 91.4 $\mathcal{J \& F}$.\\

    \noindent \textbf{DAVIS 2017}~\cite{davis17} is a multiple objects benchmark. 
    The validation set contains 59 objects in 30 videos.
    In the \cref{tab:results}, GSFM achieves an average score of 86.2, which outperforms baseline methods by 1.5 $\mathcal{J\&F}$.
    What's more, we also test our model on the more challenging DAVIS 2017 test-dev split set. It also significantly surpasses the baseline method (1.4 $\mathcal{J\&F}$).\\
    
    \noindent \textbf{YouTube-VOS}~\cite{youtube} is the largest benchmark available  for video object segmentation. 
    It contains 3471 videos in the training set (65 categories), 507 videos in the valid set (additional 26 categories not in the training set), and 541 videos in the test set.
    As shown in~\cref{youtube}, our method achieves competitive results (83.8) on YouTube-VOS and outperforms the baseline methods by 0.9 $\mathcal{J\&F}$.\\

    \begin{table}[t]
    \centering
    \caption{Evaluation on YouTube-VOS 2018 validation set. 
    Seen and Unseen denote the presence or absence of these categories in the training set, respectively. 
    $\mathcal{G}$ is the averaged score of all $\mathcal{J}$ and $\mathcal{F}$.}
    \label{youtube}
    \setlength\tabcolsep{11pt}
    \begin{tabular}{lccccc}
        \toprule[1.5pt]
        \multirow{2}{*}{Methods} &
        \multicolumn{2}{c}{Seen} & \multicolumn{2}{c}{Unseen} &
        \multirow{2}{*}{$\mathcal{G}$} \\
        \noalign{\smallskip} \cline{2-5} \noalign{\smallskip}
              & $\mathcal{J}$ & $\mathcal{F}$
              & $\mathcal{J}$ & $\mathcal{F}$ \\
        \midrule
    OnAVOS~\cite{onavos}    & 60.1    & 62.7  & 46.6  & 51.4  & 55.2         \\
    PReMVOS~\cite{premvos}  & 71.4    & 75.9  & 56.5    & 63.7  & 66.9          \\
    STM~\cite{stm}          & 79.7    & 84.2  & 72.8    & 80.9  & 79.4          \\
    AFB-URR~\cite{afb-urr}          & 78.8    & 83.1  & 74.1    & 82.6  & 79.6      \\
    GCM~\cite{gcm}        & 72.6    & 75.6  & 68.9    & 75.7  & 73.2          \\
    KMN~\cite{kmn}          & 81.4    & 85.6  & 75.3   & 83.3  & 81.4        \\
    G-FRTM~\cite{g-frtm}    & 68.6    & 71.3  & 58.4   & 64.5  & 65.7       \\
    SwiftNet~\cite{swiftnet} & 77.8   & 81.8  & 72.3   & 79.5  & 77.8         \\
    SSTVOS~\cite{sstvos}     & 80.9   & -     & 76.6     & -     & 81.8     \\
    RMNet~\cite{rmnet}       & 82.1   & 85.7  & 75.7   & 82.4  & 81.5        \\
    LCM~\cite{lcm}           & 82.2 & 86.7 & 75.7  & 83.4    & 82.0     \\
    MiVOS~\cite{mivos}       & 80.0    & 84.6 & 74.8   & 82.4  & 80.4        \\
    JOINT~\cite{joint}                 & 81.5 & 85.9 & 78.7 & 86.5 & 83.1 \\
    HMMN~\cite{hmm}                 & 82.1 & 87.0 & 76.8 & 84.6 & 82.6 \\
    RPCMVOS~\cite{aaai} &83.1 &87.7 &78.5 &86.7 &84.0 \\
    DMN-AOA~\cite{alignment}                 & 82.5 & 86.9 & 76.2 & 84.2 & 82.5 \\
    AOT-L~\cite{aot}  & 82.5   & 87.5 & 77.9 & 86.7 & 83.7 \\
    STCN*~\cite{stcn}                 & \underline{81.8} & \underline{86.4} & \underline{77.8} & \underline{85.6} & \underline{82.9} \\
    \midrule
    \textbf{GSFM (Ours)}      & \textbf{82.8}  & \textbf{87.5}  & \textbf{78.3}   & \textbf{86.5} & \textbf{83.8}\\
    \bottomrule[1.5pt]
    \end{tabular}
    \end{table}

\begin{figure*}[t]
    \centering
    \includegraphics[width=\textwidth]{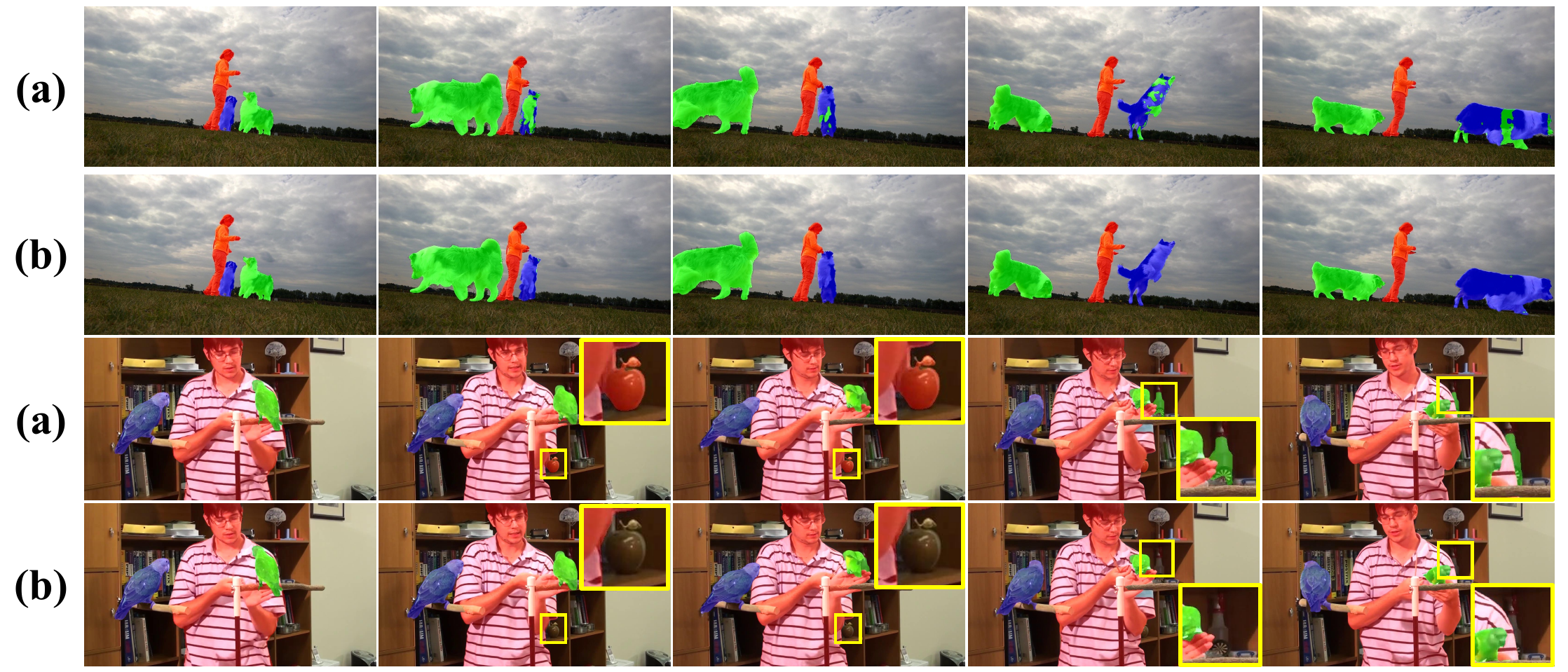}
    \caption{Visualization results of our proposed method. (a) denotes the segmentation results of our baseline method~\cite{stcn}. (b) is the results of our GSFM. The first example shows that our model can better perceive the overall semantic information of the target and thus identify similar objects. The second example shows that our approach makes a better determination of the ambiguous areas}
    \label{visualization}
\end{figure*}

\noindent \textbf{Qualitative Results.}
\cref{visualization} shows some comparison examples between ours GSFM and STCN~\cite{stcn}.
In the first example, similar pixels of the dogs are easily mis-segmented by STCN because only local information is used for matching.
While with LFM enhancing global semantic information, GSFM can identify targets more robustly.
This is also illustrated in~\cref{intro}.
The second example shows that with HFM enhancing fine-grained information, the proposed model has a better judgment for details and ambiguous areas.

\subsection{Ablation Study}

\begin{minipage}{\textwidth}
 \begin{minipage}[h]{0.45\textwidth}
  \centering
     \makeatletter\def\@captype{table}\makeatother
     \caption{Enhancing different frequency. $\text{freq}_{L}$, $\text{freq}_{H}$, and $\text{freq}_{F}$ denotes enhancing low, high, and full-frequency, respectively}
     \label{tab:lfm}
     \small
     \renewcommand\tabcolsep{4pt}
           \begin{tabular}{ccccc}
    \toprule[1.5pt]
    LFM &HFM   & $\mathcal{J}$ & $\mathcal{F}$ & $\mathcal{J}\&\mathcal{F}$ \\ 
    \midrule
    $\text{freq}_{F}$ & $\text{freq}_{H}$    & 81.6  &  88.0 & 84.8$^{\downarrow0.5}$   \\ 
    $\text{freq}_{H}$ &$\text{freq}_{H}$    &80.8  & 87.6 &84.2$^{\downarrow1.1}$ \\
    Attn. & $\text{freq}_{H}$   &81.6 & 88.2   & 84.9$^{\downarrow0.4}$    \\\midrule
    $\text{freq}_{L}$ & $\text{freq}_{F}$     &81.4  &  87.9     & 84.7$^{\downarrow0.6}$   \\ 
    $\text{freq}_{L}$ &$\text{freq}_{L}$    &81.2  & 87.7  &84.5$^{\downarrow0.8}$ \\
    \midrule
    \textbf{$\text{freq}_{L}$} & \textbf{$\text{freq}_{H}$} & \textbf{81.9} & \textbf{88.7} &\textbf{85.3} \\
    \bottomrule[1.5pt]
    \end{tabular}
  \end{minipage}
  \begin{minipage}[h]{0.05\textwidth}
  \ 
  \end{minipage}
  \begin{minipage}[h]{0.43\textwidth}
  \centering
        \makeatletter\def\@captype{table}\makeatother\caption{The quantitative results of generalization effect. FM denotes the proposed LFM and HFM and \checkmark \ indicates deployed}
        \small
        \label{plug}
        \renewcommand\tabcolsep{3pt}
    \begin{tabular}{lcccc}
    \toprule[1.5pt]
                    Method    &  FM & $\mathcal{J}$             & $\mathcal{F}$      & $\mathcal{J}\&\mathcal{F}$       \\ \midrule
    \multirow{2}{*}{STM~\cite{stm}}         &            & 78.8          & 84.2    & 81.5    \\      
                                 &\checkmark  & 80.8 & 86.2 & \textbf{83.5}$^{\uparrow}$\\ \midrule
    \multirow{2}{*}{KMN~\cite{kmn}}         &          & 79.7          & 85.5     & 82.6      \\   
                                 &\checkmark  & 81.6 & 87.8 & \textbf{84.7}$^{\uparrow}$ \\ \midrule
    \multirow{2}{*}{MiVOS~\cite{mivos}}       &           & 79.8          & 85.6    & 82.7    \\   
                                 &\checkmark  & 81.7 & 87.4 & \textbf{84.6}$^{\uparrow}$\\
    \bottomrule[1.5pt]
    \end{tabular}
   \end{minipage}
\end{minipage}

\subsubsection{{Analysis on LFM and HFM.}}
In addition to the observation in some theoretical works ~\cite{frequency-analysis2}, we conduct experiments to verify the rationality of enhancing low-frequency in encoder and high-frequency in decoder.
The results are shown in \cref{tab:lfm}.
Note that enhancing full frequency is different from removing the module since it still updates the features in the spectral domain.
From the table we can see that, when the high-frequency components are enhanced in the encoder, there is a significant decrease on performance (1.1 $\mathcal{J\&F}$), which illustrates the encoded features need contain enough high-level semantic information.
Conversely, 
decoder features need have fine-grained detail information.
Besides, we have also tried other strategy that fusing global information, \textit{e.g.}, attention, and LFM works better.\\

\noindent \textbf{Generalizability Analysis.}
To demonstrate the generalization ability of our frequency modules and prove that the lack of intra-frame global dependency is a common problem of memory network-based methods, we conduct experiments by applying our modules on some other methods as well.
As shown in~\cref{plug}, the effectiveness of these methods is significantly improved by adding the frequency modules, which further shows the rationality of enhancing different frequency components separately in different parts of the network.\\

\begin{figure}[t]
\centering  
\subfigure[Quantitative results]{   
\begin{minipage}[t]{0.45\textwidth}
\centering    
\label{cutoff1}
\includegraphics[width=\linewidth]{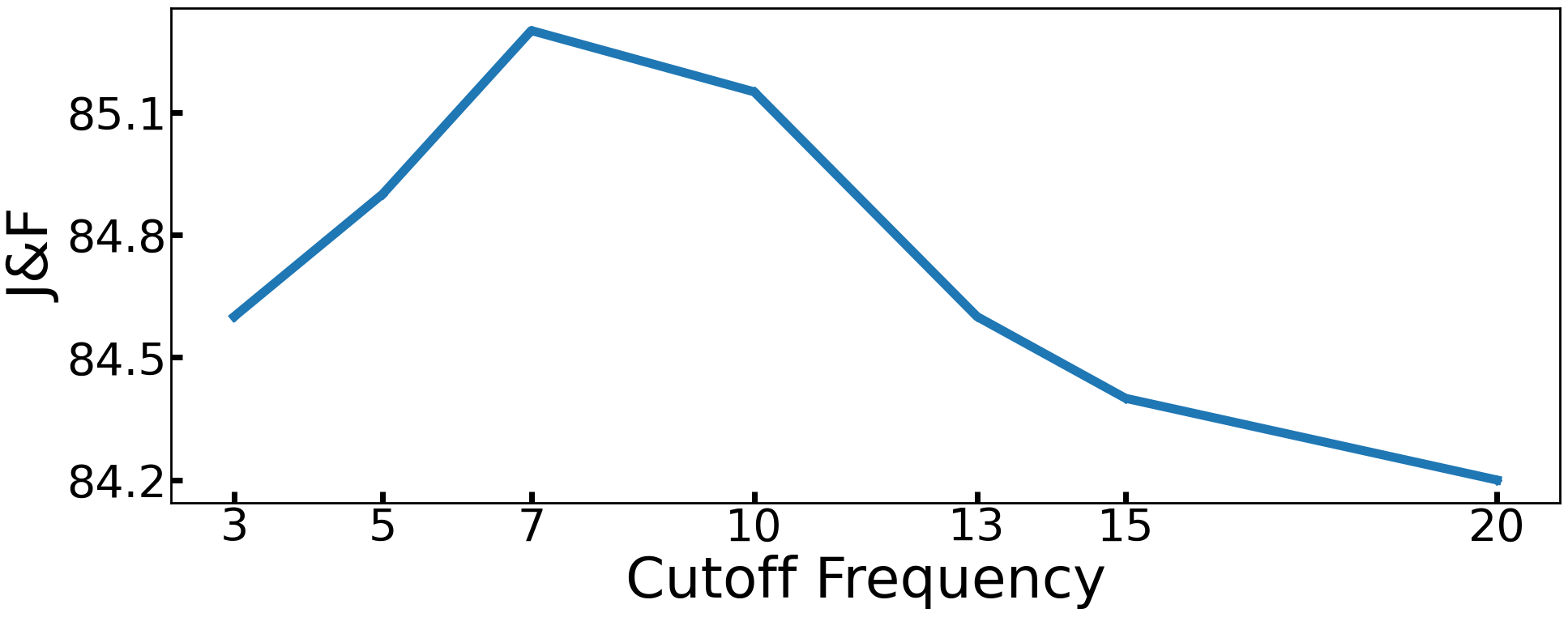}  
\end{minipage}
}
\subfigure[Visualization of filter kernels]{ 
\begin{minipage}[t]{0.45\textwidth}
\centering    
\label{cutoff2}
\includegraphics[width=\linewidth]{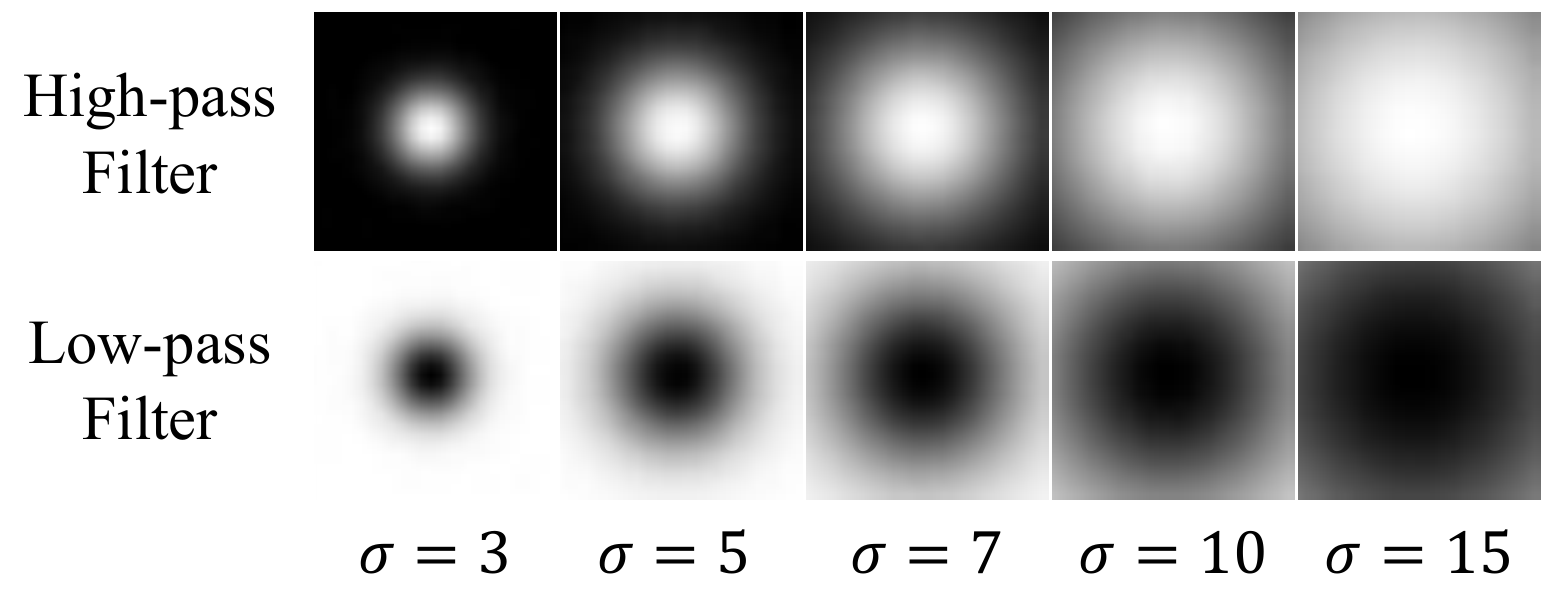}
\end{minipage}
}
\caption{Analysis on the selection of frequency filters}    
\label{cutoff}    
\end{figure}

\noindent \textbf{Selection of Frequency Filters.}
When performing frequency enhancement, we need to choose the cutoff frequency $\sigma$ that distinguishes high and low frequency (the value of the cutoff frequency affects the frequency filter).
After visualizing and experimenting with Gaussian filter kernels of different cutoff frequencies, finally, we choose $\sigma = 7$ as the cutoff frequency in default.
\cref{cutoff1} shows that too large or too small cutoff frequency will have a bad effect.
From~\cref{cutoff2} we can see that if $\sigma$ is set too large, the high-pass filter will pass almost all frequencies while the low-pass filter will filter out all frequencies, which losses the function of selective enhancement.
Same thing if $\sigma$ is set too small.\\

\noindent \textbf{Component Analysis.}
We experiment with the effectiveness of the proposed LFM, HFM, and Boundary Decoder.
As shown in Table~\ref{component analysis}, all of them bring performance improvement and their combination works better (upgraded 1.2).\\

    \begin{table}[t]
    \centering
    \caption{Ablation study of the proposed modules}
    \label{component analysis}
    \renewcommand\tabcolsep{10pt}
    \begin{tabular}{ccccccc}
    \toprule[1.5pt]
    LFM   &HFM  &Boundary Branch  & $\mathcal{J}$ & $\mathcal{F}$ & $\mathcal{J}\&\mathcal{F}$ & FPS\\ 
    \midrule
                 &     &     & 80.8  &  87.4     & 84.1  &11.7 \\ 
    \checkmark         &        &           & 81.5  & 87.9      & \textbf{84.7}$^{\uparrow0.6}$  &11.3 \\
    &\checkmark & &81.2 &87.8 &\textbf{84.5}$^{\uparrow0.4}$ &10.9\\
    \checkmark            &     & \checkmark  & 81.8 & 88.2   & \textbf{85.0}$^{\uparrow0.9}$   &9.5 \\
             & \checkmark  &\checkmark  & 81.4 & 87.7   & \textbf{84.6}$^{\uparrow0.5}$    &9.1\\
    \checkmark        & \checkmark         & \checkmark & 81.9 & 88.7  & \textbf{85.3}$^{\uparrow1.2}$  &8.9 \\
    \bottomrule[1.5pt]
    \end{tabular}
    \end{table}

\begin{figure}[t]
    \centering
    \begin{minipage}[b]{0.54\linewidth}
        \includegraphics[width=1.0\linewidth]{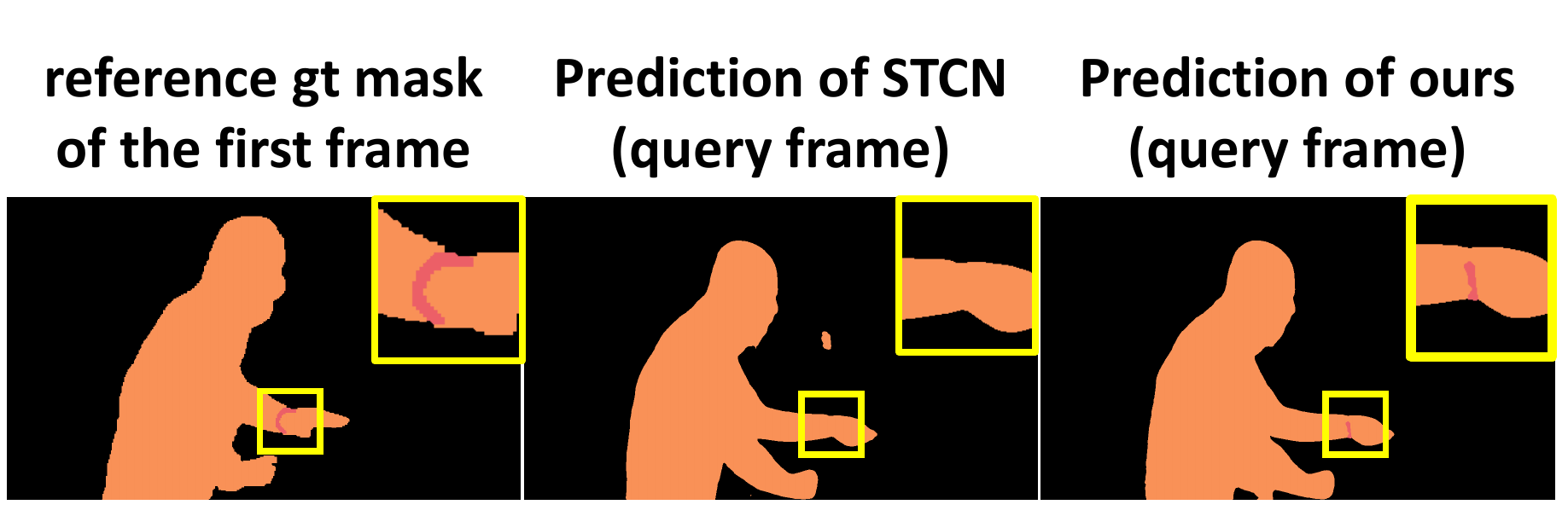}
        \caption{Qualitative results on small objects.}
        \label{fig:small_object}
    \end{minipage}
    \hfill
    \begin{minipage}[b]{0.44\linewidth}
    \centering
    \small
    \setlength{\tabcolsep}{8pt}
        \begin{tabular}{lccc}
            \toprule
            Area     & 5\%   & 1\% & 0.5\% \\
            \hline
            STCN          &  80.6  &  76.3 & 73.5   \\
            Ours       &  81.4  &  78.1  &75.0 \\
            \bottomrule
        \end{tabular}
        \captionof{table}{Quantitative results on small objects.}
        \label{tab:small_object}
    \end{minipage}
\end{figure}

\noindent \textbf{Effect of Small Objects.}
Although the LFM takes a residual structure to enhance low-frequency components during encoding, it does not result in information loss.
To prove that, we analyze the segmentation effect of small objects on YouTubeVOS dataset. \cref{fig:small_object} and \cref{tab:small_object} show the qualitative results and quantitative results respectively.
In \cref{tab:small_object}, we count the results for objects with area less than 5\%, 1\% and 0.5\% of the image.
It can be seen that the segmentation results of small objects are not worse, but better due to the enhanced discrimination of features. 

\section{Conclusions}
To fully utilize the intra-frame spatial dependency, we propose a Global Spectral Filter Memory network (GSFM) for semi-supervised video object segmentation in this paper.
According to the different characteristics of encoding and decoding, GSFM separately enhances corresponding frequency components.
With LFM integrating high-level semantic information and HFM highlighting fine-grained details, GSFM shows excellent performance on the popular DAVIS~\cite{davis16,davis17} and YouTube-VOS~\cite{youtube}.
Besides, extensive experiments also demonstrate the rationality and generalization ability of our frequency modules.
We hope that the strategy enhancing low-frequency for encoding and high-frequency for decoding would inspire some research in related fields.

\subsection*{Acknowledgement.}
This work was partially supported by the National Natural Science Foundation of China under Grant No. U1903213 and the Shenzhen Key Laboratory of Marine IntelliSense and Computation (NO. ZDSYS20200811142605016.)

\clearpage
%
%
\bibliographystyle{splncs04}
\bibliography{egbib}
\end{document}